\documentclass[letterpaper]{article} 
\usepackage{aaai24}  
\usepackage{times}  
\usepackage{helvet}  
\usepackage{courier}  
\usepackage[hyphens]{url}  
\usepackage{graphicx} 
\usepackage{natbib}  
\usepackage{caption} 
\usepackage{makecell}
\usepackage{booktabs}
\usepackage{amsmath}
\usepackage{amsfonts}
\usepackage{tabularx}
\usepackage{pdfpages} 
\usepackage{pgffor} 

\makeatletter
\AtBeginDocument{\let\LS@rot\@undefined}
\makeatother

\def\supplementfilename{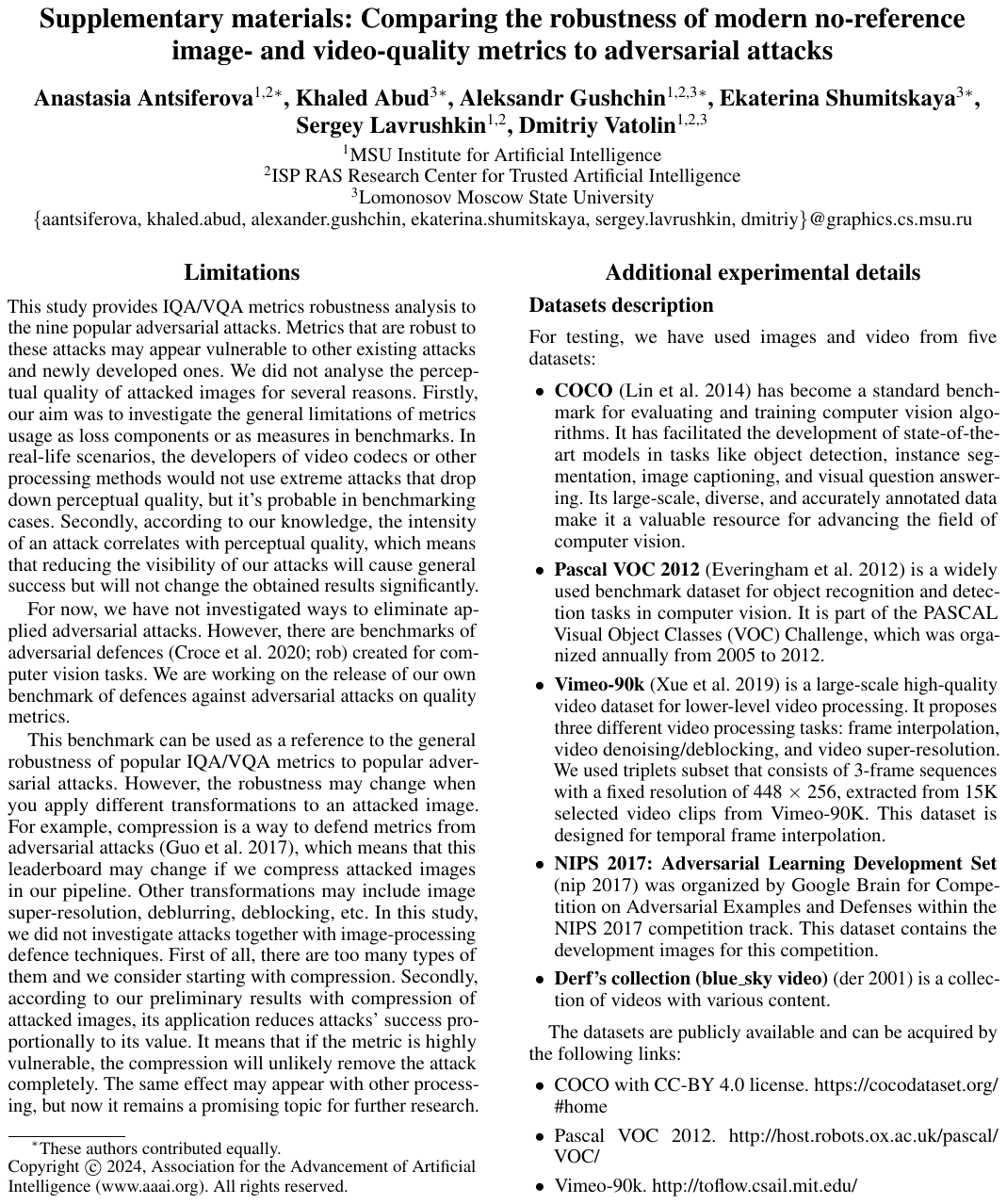}

\pdfximage{\supplementfilename}
\def\numbersupplementpages{\the\pdflastximagepages}

\newif\ifarXiv
\arXivtrue 

\usepackage{algorithm}
\usepackage{algorithmic}

\usepackage{newfloat}
\usepackage{listings}
\DeclareCaptionStyle{ruled}{labelfont=normalfont,labelsep=colon,strut=off} 
\floatstyle{ruled}
\newfloat{listing}{tb}{lst}{}
\floatname{listing}{Listing}
\title{Comparing the Robustness of Modern No-Reference Image- and Video-Quality Metrics to Adversarial Attacks}

\author {
    Anastasia Antsiferova\textsuperscript{\rm 1,2},
    Khaled Abud\textsuperscript{\rm 3},
    Aleksandr Gushchin\textsuperscript{\rm 1,2,3},
    Ekaterina Shumitskaya\textsuperscript{\rm 3},
    Sergey Lavrushkin\textsuperscript{\rm 1,2},
    Dmitriy Vatolin\textsuperscript{\rm 1,2,3}
}
\affiliations {
    \textsuperscript{\rm 1}MSU Institute for Artificial Intelligence\\
    \textsuperscript{\rm 2}ISP RAS Research Center for Trusted Artificial Intelligence\\
    \textsuperscript{\rm 3}Lomonosov Moscow State University\\
    \{aantsiferova, khaled.abud, alexander.gushchin, ekaterina.shumitskaya, sergey.lavrushkin, dmitriy\}@graphics.cs.msu.ru
}

\begin{document}

\maketitle

\begin{abstract}
Nowadays, neural-network-based image- and video-quality metrics perform better than traditional methods. However, they also became more vulnerable to adversarial attacks that increase metrics' scores without improving visual quality. The existing benchmarks of quality metrics compare their performance in terms of correlation with subjective quality and calculation time. Nonetheless, the adversarial robustness of image-quality metrics is also an area worth researching. This paper analyses modern metrics' robustness to different adversarial attacks. We adapted adversarial attacks from computer vision tasks and compared attacks' efficiency against 15 no-reference image- and video-quality metrics. Some metrics showed high resistance to adversarial attacks, which makes their usage in benchmarks safer than vulnerable metrics. The benchmark accepts submissions of new metrics for researchers who want to make their metrics more robust to attacks or to find such metrics for their needs. The latest results can be found online: \url{https://videoprocessing.ai/benchmarks/metrics-robustness.html}.
\end{abstract}

\section{Introduction}
\label{intro}

Nowadays, most new image- and video-quality metrics (IQA/VQA) employ deep learning. For example, in the latest NTIRE challenge on perceptual quality assessment ~\cite{gu2022ntire}, all winning methods were based on neural networks. With the increased sizes of datasets and availability of crowdsourced markup, deep-learning-based metrics started to outperform traditional approaches in correlation with subjective quality. However, learning-based methods, including IQA/VQA metrics, are more vulnerable to adversarial attacks. A simple metric like PSNR is more stable to image modifications that aim to manipulate quality scores (any changed pixel will decrease the score). In contrast, the behaviour of deep metrics is much more complex. The existing benchmarks evaluate metrics' correlation with subjective quality but do not consider their robustness. At the same time, \textit{the possibility to manipulate IQA/VQA metrics scores is already being exploited in different real-life scenarios}. Below are some examples of such scenarios and potential negative impacts from using non-robust IQA/VQA.


\textit{Decrease of perceptual quality.} Metrics-oriented optimization modes are already being implemented in video encoders. libaom \cite{deng2020vmaf} and LCEVC \cite{lcevc_tune_vmaf} have options that optimize bitstream for increasing a VMAF score. Such tuning was designed to improve the visual quality of the encoded video; however, as VMAF is a learning-based metric, it may decrease perceptual quality \cite{zvezdakova2019hacking, siniukov2021hacking}. 
Using unstable image quality metrics as a perceptual proxy in a loss function may lead to incorrect restoration results \cite{ding2021comparison}. For instance, LPIPS is widely used as a perceptual metric, but optimizing its scores leads to increased brightness \cite{kettunen2019lpips}, which is unwanted or even harmful (for example, when analyzing medical images). 
    
\textit{Cheating in benchmarks}. The developers of image- and video-processing methods can use metrics' vulnerabilities to achieve better competition results. For example, despite LPIPS already being shown to be vulnerable to adversarial attacks, it is still used as the main metric in some benchmarks, e.g. to compare super-resolution methods \cite{zhang2021benchmarking}. 
In some competitions that publish the results of subjective comparisons and objective quality scores, we can see the vast difference in these leaderboards. For instance, the VMAF leaders in 2021 Subjective Video Codecs Comparisons differ from leaders by subjective quality \cite{msu_subjective_2021}.
    
\textit{Manipulating the results of image web search.} Search engines use not only keywords and descriptions but also image quality measurement to rank image search results. For example, the developers of Microsoft Bing used image quality as one of the features to improve its output \cite{microsoft_bing_2013}. As shown in MediaEval 2020 Pixel Privacy: Quality Camouflage for Social Images competition \cite{pixel_privacy_2020}, there are a variety of ways to fool image quality estimators. 

Our study highlights the necessity of measuring the adversarial robustness of contemporary metrics for the research community. There are different ways to cheat on IQA/VQA metrics, such as increasing or decreasing their scores. In our study, we focus on analyzing metrics' resistance to attacks that increase estimated quality scores, as this kind of attack has already appeared in many real-life cases. Also, by choosing to investigate metrics' stability to scores increasing, we do not limit the generability of the results. We believe that the existing image- and video-quality metrics benchmarks must be supplemented with metrics' robustness analysis. In this paper, we first attempt to do this and apply several types of adversarial attacks to a number of quality metrics. Our contributions are as follows: a new benchmark methodology, a leaderboard published online \footnote{\url{https://videoprocessing.ai/benchmarks/metrics-robustness.html}}, and an analysis of currently obtained results. We published our code \footnote{\url{https://github.com/msu-video-group/MSU_Metrics_Robustness_Benchmark}} for generating adversarial attacks and a list of open datasets used in this study, so the developers of IQA/VQA methods can measure the stability of their methods to attacks. For those who want their approach published on our website, the benchmark accepts new submissions of quality metrics. Try our benchmark using \textit{pip install robustness-benchmark}.

\section{Related Work}
\label{related}

Depending on the availability of the undistorted image, IQA/VQA metrics can be divided into three types: no-reference (NR), full-reference (FR) or reduced-reference (RR). NR metrics have the broadest applications but generally show lower correlations with subjective quality than FR and RR metrics. However, recent results show that new NR metrics outperformed many existing FR methods, so we mainly focused on NR metric evaluation in this paper. The performance of IQA/VQA metrics is traditionally evaluated using subjective tests that measure the correlation of metric scores with perceptual ones. The most well-known comparisons were published within NTIRE Workshop \cite{gu2022ntire}, and two benchmarks currently accept new submissions: MSU Video Quality Metrics Benchmark \cite{NEURIPS2022_59ac9f01} and UGC-VQA \cite{tu2021video}. These studies show how well the compared metrics estimate subjective quality but do not reflect their robustness to adversarial attacks.

There are different ways to measure the robustness of neural network-based methods. It can be done via theoretical estimations, e.g. Lipschitz regularity. However, this approach has many limitations, including the number of parameters in the evaluated network. A more universal approach is based on applying adversarial attacks. This area is widely studied for computer vision models. However, not all methods can be adapted to attack quality metrics. 

The first methods for measuring the robustness of IQA/VQA metrics were based on creating a specific situation in which the metric potentially fails. Ciaramello and Reibman \shortcite{ciaramello2011supplemental} first conducted such analysis and proposed a method to reveal the potential vulnerabilities of an objective quality model based on the generation of image or video pairs with the intent to cause misclassification errors \cite{brill2004accuracy} by this model. Misclassification errors include false ordering (FO, the objective model rates a pair opposite to humans), false differentiation (FD, the objective model rates a pair as different but humans do not), and false tie (FT, humans order a pair as different, but the objective model does not). 
H. Liu and A. Reibman \shortcite{liu2016software} introduced a software called ``STIQE'' that automatically explores an image-quality metric's performance. It allows users to execute tests and then generate reports to determine how well the metric performs. Testing consists of applying several varying distortions to images and checking whether the metric score rises monotonically as the degree of the applied distortion.

Nowadays, metrics' adversarial robustness is primarily estimated by adapting attacks designed for computer vision tasks to image quality metrics. 
A more detailed description of existing attacks against metrics that we used in our study is given in the section ``List of adversarial attacks''. There are two recently published attacks that we aim to add to the benchmark shortly: a new CNN-based generative attack FACPA \cite{shumitskaya2023fast}, attack with human-in-the-loop by Zhang et al. \cite{zhang2022perceptual} and spatial attack that was adapted for metrics \cite{ghildyal2023attacking}.

Recently, a new study on the adversarial robustness of full-reference metrics was published \cite{ghildyal2023attacking}. The authors showed that six full-reference metrics are susceptible to imperceptible perturbations generated via common adversarial attacks such as FGSM \cite{DBLP:journals/corr/GoodfellowSS14}, PGD \cite{madry2017towards}, and the One-pixel attack \cite{su2019one}. They also showed that adversarial perturbations crafted for LPIPS metric \cite{zhang2018unreasonable} using stAdv attack can be transferred to other metrics. As a result, they concluded that more accurate learning-based metrics are less robust to adversarial attacks than traditional ones. 
We summarised the existing research on IQA/VQA metrics' robustness to adversarial attacks in Table~\ref{tab:related_work}.

\begin{table}[t]
   \centering
   \begin{tabularx}{\columnwidth}{ccccc}
     \hline 
     Benchmark & \makecell{\# attacks / \\ \# metrics} & \makecell{Metrics \\ type} & \makecell{Test \\datasets} \\
     \hline
     \makecell{Ciaramello and \\ Reibman \shortcite{ciaramello2011supplemental}} & 5 / 4 & FR & 10 images\\
      \makecell{Ciaramello and \\ Reibman \shortcite{ciaramello2011systematic}} & 5 / 9 & NR, FR & 473 images \\
      \makecell{Liu and \\ Reibman \shortcite{liu2016software}} & 5 / 11 & NR, FR & 60 images \\
      \makecell{Shumitskaya \\ et al. \shortcite{Shumitskaya_2022_BMVC}} & 1 / 7 & NR & 20 videos \\
      Zhang et al. \shortcite{zhang2022perceptual} & 1 / 4 & NR & 12 images \\
      \makecell{Ghildyal and \\ Liu \shortcite{ghildyal2023attacking}} & 6 / 5 & FR & \makecell{12,227 \\ images} \\
      \hline
      Ours & 9 / 15 & NR, FR & \makecell{3000 \\ images, \\1 video} \\
     \hline
   \end{tabularx}
   \caption{Comparisons of image- and video-quality metrics' stability to adversarial attacks.}
   \label{tab:related_work}
 \end{table}

\section{Benchmark}
\label{benchmark}

\subsection{List of Metrics}

In this paper, we focused on the evaluation of only no-reference metrics for several reasons: firstly, there exists a similar evaluation of full-reference metrics \cite{ghildyal2023attacking}; secondly, no-reference metrics have a more comprehensive range of applications and are more vulnerable to attacks; thirdly, these metrics are mostly learning-based. We considered state-of-the-art metrics according to other benchmarks and various other no-reference metrics. All tested metrics assess image quality, except for VSFA \cite{li2019quality} and MDTVSFA \cite{li2021unified}, which are designed for videos. 

\textbf{RankIQA} \cite{liu2017rankiqa} pre-trains a model on a large dataset with synthetic distortions to compare pairs of images, then fine-tunes it on a small realistic dataset. \textbf{MetaIQA} \cite{zhu2020metaiqa} introduces a quality prior model pre-trained on several dozens of specific distortions and fine-tuned on a smaller target dataset, similar to RankIQA. \textbf{WSP} \cite{su2020blind} is concerned with Global Average Pooling feature aggregation used by most existing methods and replaces it with Weighted Spatial Pooling to distinguish important locations. \textbf{CLIP-IQA} \cite{wang2022exploring} predicts the quality perception and image-provoked abstract emotions by feeding heterogeneous text prompts and the image to the CLIP network. \textbf{PAQ-2-PIQ} \cite{ying2020patches} introduces a large subjective picture quality database of about 40,000 images, trains a CNN with ResNet-18 backbone to predict patch quality and combines the predictions with RoI pooling. \textbf{HyperIQA} \cite{su2020blindly} focuses on real-life IQA and proposes a hyperconvolutional network that predicts the weights of fully connected layers. \textbf{MANIQA} \cite{yang2022maniqa} assesses quality of GAN-based distortions. The model uses vision transformer features processed by proposed network modules to enhance global and local interactions. The final score prediction utilizes patch weighting. \textbf{TReS} \cite{golestaneh2022no} proposes to compute local features with CNN and non-local features with self-attention, introduces a per-batch loss for correct ranking and a self-supervision loss between reference and flipped images. \textbf{FPR} \cite{chen2022no} hallucinates pseudo-reference features from the distorted image using mutual learning on reference and distorted images with triplet loss. Attention maps are predicted to aggregate scores over patches. \textbf{VSFA} \cite{li2019quality} estimates video quality using ResNet-50 features for content awareness and differentiable temporal aggregation, which consists of gated recurrent units with min pooling. \textbf{MDTVSFA} \cite{li2021unified} enhances VSFA with explicit mapping between predicted and dataset-specific scores, supported by multi-dataset training. \textbf{NIMA} \cite{talebi2018nima}  predicts a distribution of scores instead of regressing a single value and considers both technical and aesthetic image scores. It is trained on the Aesthetic Visual Analysis database using squared earth mover's distance as a loss. \textbf{LINEARITY} \cite{li2020norm} invents the norm-in-norm loss, which shows ten times faster convergence than MSE or MAE with ResNet architecture. \textbf{SPAQ} \cite{fang2020perceptual} collects a database of 11,125 smartphone photos, proposes a ResNet-50 baseline model and three modified versions incorporating EXIF data (MT-E), subjective image attributes (MT-A) and scene labels (MT-S). \textbf{KonCept512} \cite{hosu2020koniq} collects KonIQ-10k, a diverse crowdsourced database of 10,073 images and trains a model with InceptionResNetV2 backbone.

We also used MSE, PSNR and SSIM \cite{ssim} as proxy metrics to estimate image quality degradation after attacks. The choice is motivated by their structure (full-reference and not learning-based), which makes them more stable to adversarial attacks.




\subsection{List of Adversarial Attacks}
\label{attacks}
In all attacks, we define the loss function as $J(\theta, I) = 1-score(I)/range$ and minimize it by making small steps along the gradient direction in image space, which increases the attacked metric score. $range$ is computed as the difference between maximum and minimum metric values on the dataset and serves to normalize the gradient magnitude across different metrics.

\textbf{FGSM-based attacks} are performed for each image. The pixel difference is limited by $\varepsilon$.
\textbf {FGSM} \cite{DBLP:journals/corr/GoodfellowSS14} is a basic approach that makes one gradient step: $I^{adv} = I - \varepsilon \cdot sign(\nabla_I J(\theta, I ))$.
\textbf {I-FGSM} \cite{kurakin2018adversarial} is a more computationally expensive method that uses $T$ iterations and clips the image on each step: $I_{t+1}^{adv} = Clip_{I, \varepsilon}\{I_t^{adv} - \alpha \cdot sign(\nabla_I J(\theta, I_t^{adv}))\}$,
where $t=0,1,\ldots T-1$, $I_0$ is the input image $I$, and $\alpha$ is the perturbation intensity. The clipped pixel value at position $(x, y)$ and channel $c$ satisfies \( |I^{adv}_t(x, y, c) - I(x, y, c)| < \varepsilon \) .
PGD \cite{madry2017towards} is identical to I-FGSM except for the random initialization in the $\varepsilon$-vicinity of the original image; due to its similarity to I-FGSM, we didn't include it in the experiments.
\textbf {MI-FGSM} \cite{dong2018boosting} uses gradient momentum: \( I_{t+1}^{adv} = Clip_{I, \varepsilon}\{I_t^{adv} - \alpha \cdot sign(g_t)\},\; t=0,1,\ldots T-1 \),\\ \( g_t = \nabla_I J(\theta,\ I_t^{adv}) + \nu \cdot g_{t-1},\;\; g_{-1}=0 \), where $\nu$ controls the momentum preservation.
\textbf {AMI-FGSM} \cite{sang4112969generation} is identical to MI-FGSM, except the pixel difference limit $\varepsilon$ is set to $1 / NIQE(I)$ by computing the NIQE \cite{mittal2012making} no-reference metric.

\textbf{Universal Adversarial Perturbation (UAP)-based attacks} generate adversarial perturbation for an attacked metric, which is the same for all images and videos. When UAP is generated, the attack process consists of the mere addition of an image with UAP. The outcome is the image with an increased target metric score. We used three methods to train UAPs.
\textbf{Cumulative-UAP} is obtained by averaging non-universal perturbation on the training dataset. Non-universal perturbations are generated using one step of gradient descent. 
\textbf{Optimized-UAP} is obtained by training UAP weights using batch training with Adam optimizer and loss function defined as target metric with opposite sign. 
\textbf{Generative-UAP} is obtained by auxiliary U-Net generator training. The network is trained to generate a UAP from random noise with uniform distribution. The Adam optimizer is used for training, and the loss function is defined as the target metric with the opposite sign. Once the network is trained, a generated UAP is saved and further used to attack new images.

\textbf{Perceptual-aware attacks} use other image quality metrics to control attack imperceptibility to the human eye.
\textbf{Korhonen et al.} \cite{korhonen2022adversarial} proposes a method for generating adversarial images for NR quality metrics with perturbations located in textured regions. They use gradient descent with additional elementwise multiplication of gradients by a spatial activity map. The spatial activity map of an image is calculated using horizontal and vertical 3$\times$3 Sobel filters.
\textbf{MADC} \cite{wang2008maximum} is a method for comparing two image- or video-quality metrics by constructing a pair of examples that maximize or minimize the score of one metric while keeping the other fixed. In our study, we fixed MSE while maximizing an attacked metric. The projected gradient descent step and binary search are performed on each iteration. Let $g1$ be the gradient with direction to increase the attacked metric and $g2$ the gradient of MSE on some iteration. The projected gradient is then calculated as $pg = g1 - \frac{g2^T \cdot g1}{g2^T \cdot g2} \cdot g2$. After projected gradient descent, the binary search to guarantee a fixed MSE is performed (with 0.04 precision). The binary search is the process that consists of small steps along the MSE gradient: if the precision is bigger than 0.04, then steps are taken along the direction of reducing MSE and vice versa. 


\subsection{Methodology}
\subsubsection{Datasets}
\label{datasets}

This study incorporated pre-trained quality metrics as a part of our evaluation benchmark. We did not perform metrics fine-tuning on any data. We used six datasets summarised in Table \ref{tab:videodatasets}. These datasets are widely used in the computer vision field. We chose them to cover a diverse range of real-life scenarios, including images and video, with varying resolutions from $299\times299$ up to $1920\times1080$ (FullHD). 
All datasets have an open license that allows them to be used in this work. Our analysis categorized the adversarial attacks into trainable and non-trainable attacks. Three datasets were used to train adversarial attacks, and three were used for testing. We trained UAP attacks using each training dataset, resulting in three versions of each attack. These versions were subsequently evaluated on the designated testing datasets, and the results for different versions were averaged among each UAP-attack type and amplitude. Non-trainable attacks were directly evaluated on the testing datasets. We have analyzed the efficiency and generalization capabilities of both trainable and non-trainable adversarial attacks across various data domains while also considering the influence of training data on metric robustness.
NIPS 2017: Adversarial Learning Development Set \shortcite{nips-dataset} was also used to train metrics' domain transformations (described further in ``Evaluation metrics'').

\begin{table*}[htb]
   \centering
   \begin{tabular}{@{}cccc|cccc@{}}
     \toprule
     \makecell{\textbf{Training datasets} \\ \textbf{(for UAP attacks)}} & \makecell{Type} & \makecell{Number \\ of samples} & \makecell{Resolution} & \textbf{Testing datasets} & \makecell{Type} & \makecell{Number \\ of samples} & \makecell{Resolution} \\
     \midrule
     COCO \shortcite{coco-dataset} & Images & 300,000 & $640\times480$ & \makecell{NIPS \shortcite{nips-dataset}} & Images & 1,000 & $299\times299$ \\
     Pascal VOC \shortcite{pascal-voc-2012} & Image & 11,530 & $500\times333$ & \makecell{Derf's collection \\ \shortcite{derf-dataset}} & Video & \makecell{24 ($\sim 10$k \\frames)} & $1920\times1080$\\
     \makecell{Vimeo-90k \\ Train set \shortcite{vimeo-90k-dataset}} & \makecell{Triplets of \\ images} & 2,001 & $448\times256$ & \makecell{Vimeo 90k \\ Test set \shortcite{vimeo-90k-dataset}} & \makecell{Triplets of \\ images} & 11,346 & $448\times256$ \\
     \bottomrule
   \end{tabular}%
   \caption{Summary of the datasets used in our study.}
   \label{tab:videodatasets}
 \end{table*}

\subsubsection{Implementation Details}
\label{implementation_details}

We used public source code for all metrics without additional pretraining and selected the default parameters to avoid overfitting. The training and evaluation of attacks on the metrics were fully automated. We employed the CI/CD tools within a GitLab repository for our measurement procedures. We established a sophisticated end-to-end pipeline from the attacked metrics' original repositories to the resulting robustness scores to make the results entirely verifiable and reproducible. The pipeline scheme, the list of used attack's hyper-parameters and the hyperparameter choice justification are presented in the supplementary materials \cite{antsiferova2023comparing}. UAP-based attacks (UAP, cumulative UAP and generative UAP) were averaged with three different amplitudes (0.2, 0.4 and 0.8).

Quality metrics implementations were obtained from official repositories. We only modified interfaces to meet our requirements and used default parameters provided by the authors. Links to original repositories and a list of applied patches (where it was needed to enable gradients) are provided in supplementary materials \cite{antsiferova2023comparing}.  

Calculations were performed on two computers with the following characteristics:
\begin{itemize}
    \item 4 x GeForce RTX 3090 GPU, an Intel(R) Xeon(R) Gold 6226R CPU @ 2.90GHz
    \item 4 x NVIDIA RTX A6000 GPU, AMD EPYC 7532 32-Core Processor @ 2.40GHz
\end{itemize}
All calculations took a total of about 2000 GPU hours. The values of parameters ($\epsilon$, number of iterations, etc.) for the attacks are listed in the supplementary materials \cite{antsiferova2023comparing}.


\subsubsection{Evaluation Metrics}
\label{metrics}

Before calculating metrics' robustness scores, metric values are transformed with min-max scaling so that the values before the attack lie in the range [0,1].
To compensate for the nonlinear dependence between metrics \cite{zhang2022perceptual}, we converted all metrics to the same domain before comparison. MDTVSFA \cite{li2021unified} was used as the primary domain, as it shows the best correlations with MOS among tested metrics according to the MSU Video Quality Metrics benchmark results. We employed the 1-Dimensional Neural Optimal Transport \cite{korotin2023neural} method to build the nonlinear transformation between the distributions of all metrics to one general shape. We also present the results without the nonlinear transformation in the supplementary materials \cite{antsiferova2023comparing}.

\textbf{Absolute} and \textbf{Relative gain}. Absolute gain is calculated as the average difference between the metric values before and after the attack. Relative gain is the average ratio of the difference between the metric values before and after the attack to the metric value before the attack plus 1 (1 is added to avoid division problems, as values before the attack are scaled to [0,1]).
\begin{equation}
\begin{array}{c}
    Abs. gain = \frac{1}{n}\sum_{i=1}^{n}\left(f(x'_i)-f(x_i)\right),
\\
    Rel. gain = \frac{1}{n}\sum_{i=1}^{n}\frac{f(x'_i)-f(x_i)}{f(x_i) + 1},
\end{array}
\end{equation}
where $n$ is the number of images, $x_i$ is the clear image, $x'_i$ --- it's attacked counterpart, and $f(.)$ is the IQA metric function.
\newline
\textbf{Robustness score} \cite{zhang2022perceptual} $R_{score}$ is defined as the average ratio of maximum allowable change in quality prediction to actual change over all attacked images in a logarithmic scale: 
\begin{equation}
    R_{score} = \frac{1}{n}\sum_{i=1}^{n}log_{10}\left( \frac{max\{\beta_1 - f(x'_i), f(x_i) - \beta_2\}}{|f(x'_i)-f(x_i)|} \right).
\end{equation}
As metric values are scaled, we use $\beta_1=1$ and $\beta_2=0$.

\textbf{Wasserstein score} \cite{wassersteindist} $W_{score}$ and \textbf{Energy Distance score} \cite{edist} $E_{score}$ are used to evaluate the statistical differences between distributions of metric values before and after the attack. Large positive values of these scores correspond to a significant upward shift of the metric's predictions, values near zero indicate the absence of the metric’s response to the attack, and negative ones show a decrease in the metric predictions and the inefficiency of the attack. These scores are defined as corresponding distances between distributions multiplied by the sign of the difference between the mean values before and after the attack:

\begin{equation}
\begin{array}{c}
    W_{score} = W_1(\hat{P},  \hat{Q}) \cdot sign(\bar{x}_{\hat{Q}} - \bar{x}_{\hat{P}}), \\
    
    W_1(\hat{P},\hat{Q}) = \inf_{\gamma \in \Gamma(\hat{P},\hat{Q})} \int_{\mathbb{R}^2} \lvert x - y \rvert d \gamma (x, y) = \\
    
    = \int_{-\infty}^{\infty} |\hat{F}_{\hat{P}}(x) - \hat{F}_{\hat{Q}}(x)| dx;
\end{array}
\end{equation}
\begin{equation}
\begin{array}{c}
    E_{score} = E(\hat{P},  \hat{Q})\cdot sign(\bar{x}_{\hat{Q}} - \bar{x}_{\hat{P}}), \\
    E(\hat{P},\hat{Q}) = (2 \cdot \int_{-\infty}^{\infty} (\hat{F}_{\hat{P}}(x) - \hat{F}_{\hat{Q}}(x))^2 dx)^{\frac{1}{2}},
  \end{array}
\end{equation}
where \(\hat{P}\) and \(\hat{Q}\) are empirical distributions of metric values before and after the attack, \(\hat{F}_{\hat{P}}(x)\) and \(\hat{F}_{\hat{Q}}(x)\) are their respective empirical Cumulative Distribution Functions, and \(\bar{x}_{\hat{P}}\) and \(\bar{x}_{\hat{Q}}\) are their respective sample means.\\

\section{Results}
\label{results}
The main results of our study are aggregated across the different attack types, training and testing datasets. Tables and figures for other robustness measures, by specific datasets and attacks, are presented in the supplementary materials \cite{antsiferova2023comparing} and on the benchmark webpage.

\begin{figure}[tb]
  \centering
   \includegraphics[width=0.98\columnwidth]{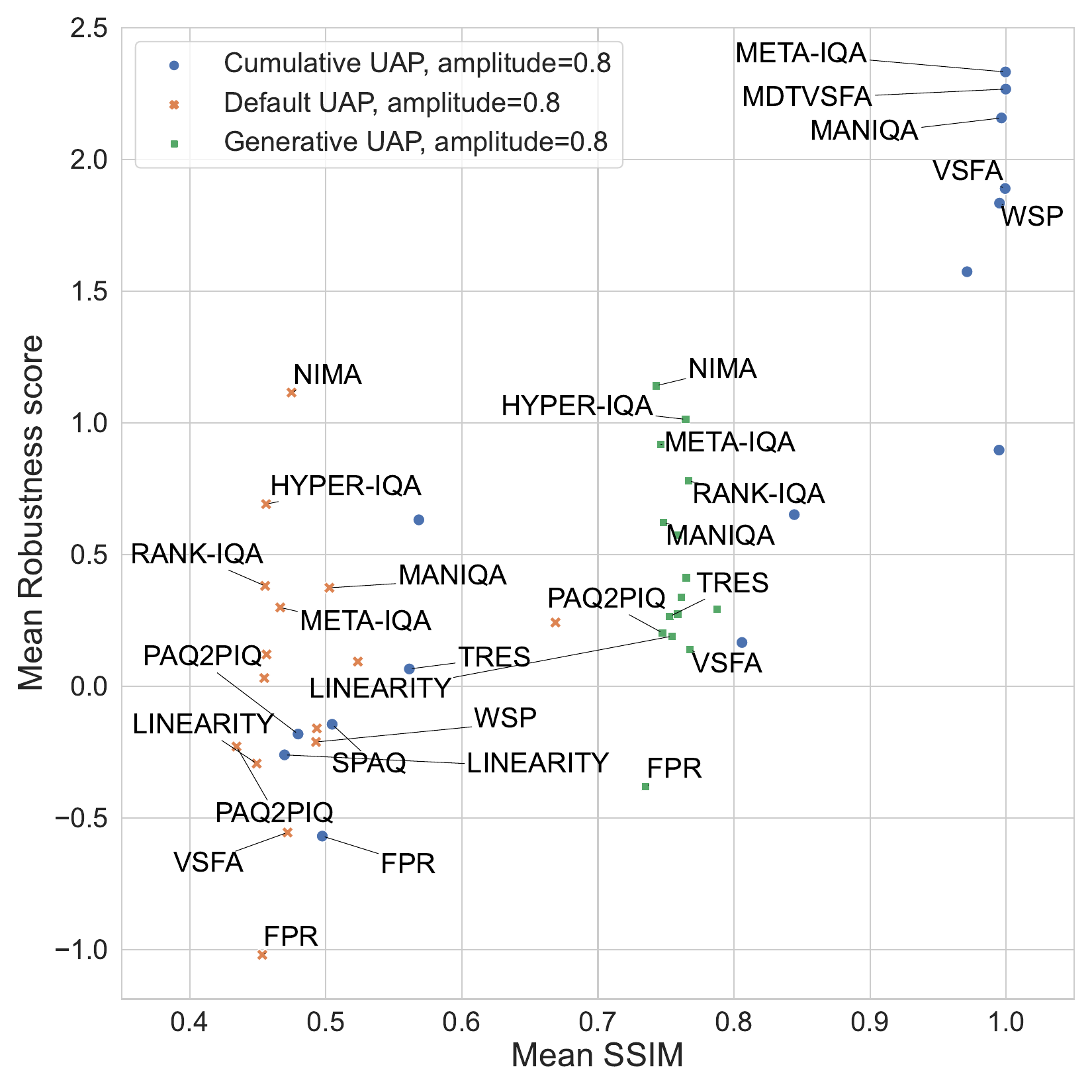}
  \caption{
  Metrics' robustness score for UAP-based adversarial attacks and SSIM measured between original and attacked images. The results are averaged for all test datasets.
}
\label{fig:uap_aggregated_mean_rscore_ssim}
\end{figure}

\textbf{Metrics that are robust to UAP-based attacks}.
Despite the three types of implemented UAP-based attacks resulting in different attack efficiency, the most and least robust metrics for these attacks are similar. MANIQA showed the best robustness score for all amplitudes of Optimized UAP and is within top-3 metrics robust to Generative UAP. This metric uses ViT and applies attention mechanisms across the channel and spatial dimensions, increasing interaction among different regions of images globally and locally. HYPER-IQA showed high resistance to all UAP attacks. Besides FPR, the PAQ-2-PIQ showed the worst energy distance score. The robustness scores of analyzed attacks are provided in Table~\ref{tab:energy_scores} and illustrated in Fig.~\ref{fig:uap_aggregated_mean_rscore_ssim}. Annotations include only five best and five worst methods judged by robustness score for better visibility.

\begin{table*}[htb]

\begin{tabularx}{\linewidth}{r|ccc|cccccc}
\toprule
    & O-UAP & G-UAP & C-UAP & FGSM & I-FGSM & MI-FGSM & AMI-FGSM & MADC & Korhonen et al. \\
\midrule
CLIP-IQA & 0.632  & 0.397 & 0.067 & 0.398 & \bfseries 0.836 & \bfseries 0.821 & \bfseries 0.819 & 0.823 & \underline {{0.812}} \\
META-IQA & 0.183 & \underline {{-0.029}} & \underline {{0.003}} & 0.529 & 1.307 & 1.285 & 1.287 & 0.934 & 0.997 \\
RANK-IQA & 0.295 & 0.064 & 0.180 & 0.285 & \underline {{1.063}} & \underline {{0.891}} & \underline {{0.893}} & \bfseries 0.383 & \bfseries 0.763 \\
HYPER-IQA & \underline {{0.072}} & \underline {{-0.094}} & 0.086 & \bfseries -0.406 & 1.366 & 1.387 & 1.396 & 0.848 & 1.329 \\
KONCEPT & 0.419 & 0.187 & 0.435 & 0.574 & 1.248 & 1.066 & 1.066 & 0.753 & 1.042 \\
FPR & 1.705 & 0.846 & 0.966 & 0.682 & 3.344 & 3.210 & 3.215 & 1.703 & 3.018 \\
NIMA & \underline {{-0.024}} & 0.046 & 0.018 & 0.258 & 1.203 & 1.147 & 1.148 & 0.959 & 1.041 \\
WSP & 0.784 & 0.155 & 0.012 & 0.405 & 1.260 & 1.251 & 1.257 & 0.760 & 0.894 \\
MDTVSFA & 0.756 & 0.359 & \underline {{0.005}} & \underline {{0.185}} & \underline {{1.011}} & \underline {{0.983}} & \underline {{0.983}} & 0.914 & \underline {{0.805}} \\
LINEARITY & 1.022 & 0.445 & 0.972 & \underline {{-0.220}} & 1.284 & 1.218 & 1.224 & 0.816 & 1.204 \\
VSFA & 1.151 & 0.361 & 0.014 & 0.306 & 2.054 & 2.272 & 2.274 & 1.470 & 1.539 \\
PAQ-2-PIQ & 0.943 & 0.252 & 0.873 & 0.578 & 1.190 & 1.123 & 1.125 & \underline {{0.536}} & 0.997 \\
SPAQ & 0.605 & 0.357 & 0.560 & 0.266 & 1.514 & 1.371 & 1.375 & 0.740 & 1.301 \\
TRES & 0.691 & 0.358 & 0.634 & 0.826 & 1.223 & 1.209 & 1.210 & 0.741 & 1.173 \\
MANIQA & \bfseries -0.390 & \bfseries -0.174 & \bfseries -0.003 & 0.499 & 1.403 & 1.225 & 1.226 & \underline {{0.698}} & 0.843 \\
\bottomrule
\end{tabularx}

\caption{Metrics' robustness calculated using energy distance score measure to different types of attacks. The results are averaged across test datasets. O-UAP stands for ``Optimised-UAP'', G-UAP for ``Generative-UAP'', C-UAP for ``Cumulative-UAP''.}
\label{tab:energy_scores}
\end{table*}

\begin{figure}[tb]
  \centering
  \includegraphics[width=0.98\columnwidth]{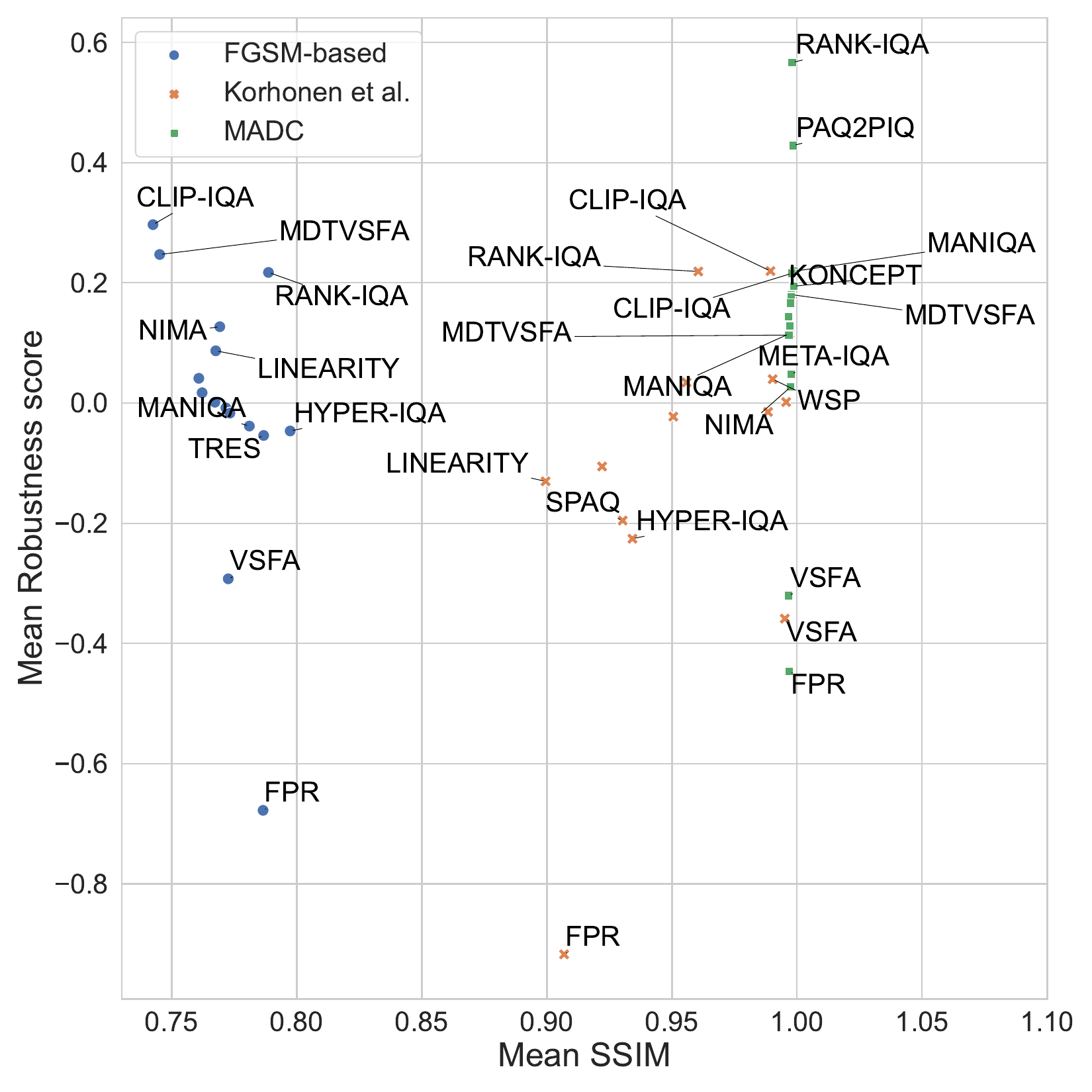}
  \caption{Metrics' robustness score for iterative adversarial attacks and SSIM measured between original and attacked images. The results are averaged for all test datasets.}
\label{fig:iterative_aggregated_mean_rscore_ssim}
\end{figure}

\textbf{Metrics that are robust to iterative attacks.}
CLIP-IQA shows the best robustness to most iterative attacks, followed by RANK-IQA and MDTVSFA. RANK-IQA also offers the best resistance to perceptually oriented MADC and Korhonen attacks. These attacks use approaches to reduce the visibility of distortions caused by an attack, which makes it more difficult for them to succeed. The robustness score of analyzed attacks is shown in Table \ref{tab:energy_scores} and illustrated in Fig.~\ref{fig:iterative_aggregated_mean_rscore_ssim}. Annotations include only five best and five worst methods judged by robustness score for better visibility.

\begin{figure}[tb]
   \centering
   \includegraphics[width=0.98\columnwidth]{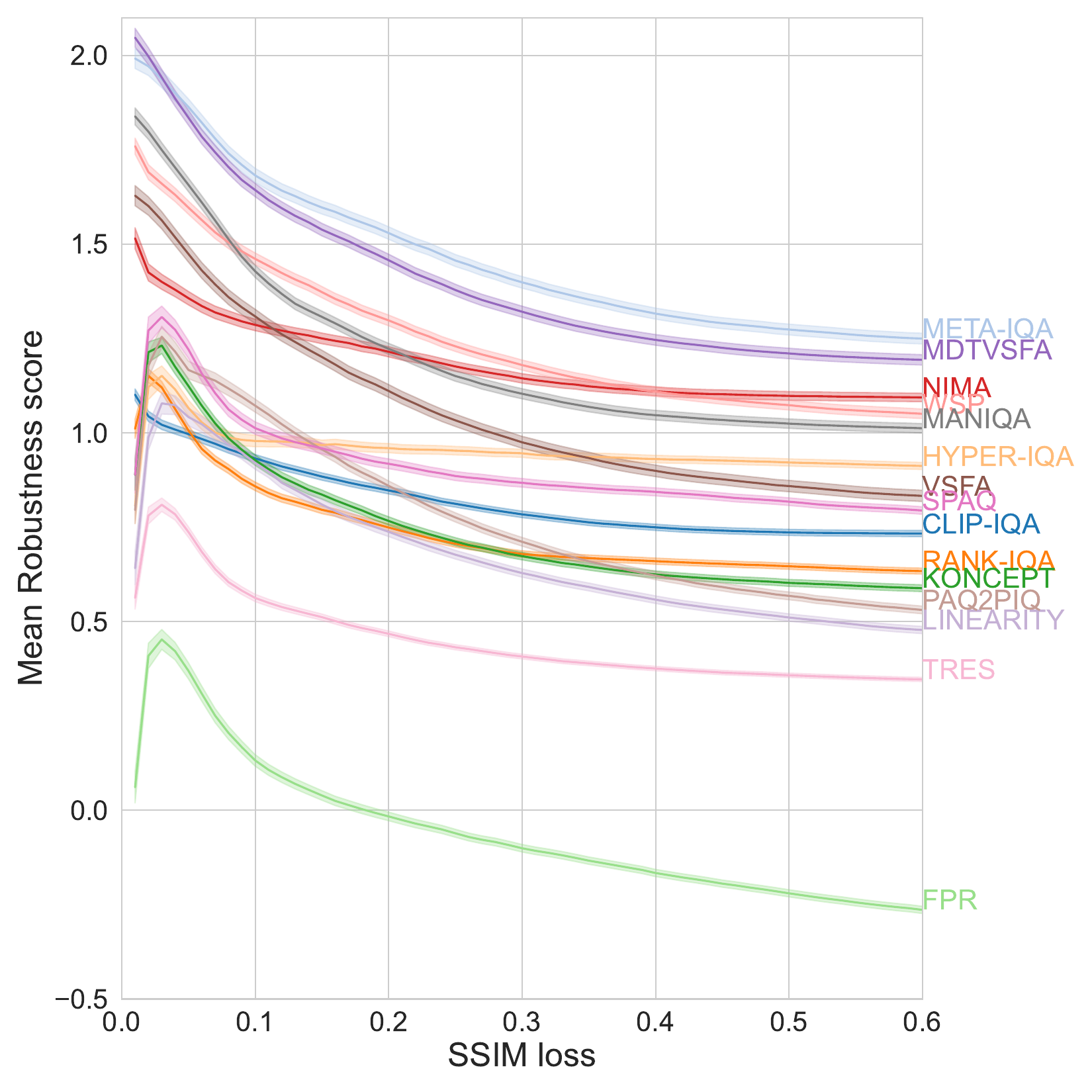}
   \caption{Dependency of metrics' robustness score of SSIM loss for attacked images (all types of attacks).}
   \label{fig:aggregated_mean_rscore}
\end{figure}

\textbf{Metrics' robustness at different levels of perceptual quality loss.}
As described in the Benchmark section, we used SSIM, PSNR and MSE as simple proxies for estimating perceptual quality loss of attacks in this study. Fig.~\ref{fig:aggregated_mean_rscore} shows an averaged robustness score depending on SSIM loss of attacked images for all attacks. It shows that all metrics become less robust to attacks when more quality degradation is allowed. HYPER-IQA's robustness is more independent from SSIM loss among all metrics. Otherwise, PAQ-2-PIQ, VSFA and FPR are becoming more vulnerable than other metrics with increasing SSIM degradation. Results for other proxy metrics (MSE and PSNR) are provided in the supplementary materials \cite{antsiferova2023comparing} and on the benchmark webpage.

\begin{table*}[tb]
\centering
\begin{tabular}{lccccc}
\toprule
 & $Abs. gain \downarrow$ & $Rel. gain \downarrow$ & $R_{score} \uparrow$ & $E_{score} \downarrow$ & $W_{score} \downarrow$ \\
\midrule
CLIP-IQA & \makecell{ 0.256 {\footnotesize(0.254, 0.258)} } & \makecell{ 0.184 {\footnotesize(0.182, 0.185)} } & \makecell{ 0.702 {\footnotesize(0.698, 0.707)} } & 0.424 & 0.256 \\
META-IQA & \makecell{ 0.241 {\footnotesize(0.238, 0.243)} } & \makecell{ 0.182 {\footnotesize(0.180, 0.184)} } & \makecell{ \textbf{ 1.168} {\footnotesize(1.161, 1.176)} } & 0.324 & 0.241 \\
RANK-IQA & \makecell{ \underline{ 0.184 } {\footnotesize(0.183, 0.186)} } & \makecell{ \underline{ 0.12 } {\footnotesize(0.119, 0.122)} } & \makecell{ 0.843 {\footnotesize(0.839, 0.848)} } & 0.285 & \underline {{0.184}} \\
HYPER-IQA & \makecell{ 0.232 {\footnotesize(0.228, 0.235)} } & \makecell{ 0.151 {\footnotesize(0.149, 0.153)} } & \makecell{ 0.740 {\footnotesize(0.735, 0.745)} } & \underline {{0.277}} & 0.237 \\
KONCEPT & \makecell{ 0.328 {\footnotesize(0.326, 0.330)} } & \makecell{ 0.227 {\footnotesize(0.225, 0.228)} } & \makecell{ 0.584 {\footnotesize(0.579, 0.589)} } & 0.489 & 0.328 \\
FPR & \makecell{ 2.591 {\footnotesize(2.568, 2.615)} } & \makecell{ 1.730 {\footnotesize(1.714, 1.746)} } & \makecell{ -0.229{\footnotesize(-0.234, -0.224)} } & 1.409 & 2.591 \\
NIMA & \makecell{ \underline{ 0.17 } {\footnotesize(0.168, 0.172)} } & \makecell{ \underline{ 0.115 } {\footnotesize(0.114, 0.117)} } & \makecell{ \underline{ 1.152 } {\footnotesize(1.146, 1.158)} } & \underline {{0.239}} & \bfseries 0.170 \\
WSP & \makecell{ 0.380 {\footnotesize(0.377, 0.384)} } & \makecell{ 0.276 {\footnotesize(0.273, 0.278)} } & \makecell{ 0.893 {\footnotesize(0.886, 0.901)} } & 0.449 & 0.380 \\
MDTVSFA & \makecell{ 0.279 {\footnotesize(0.277, 0.281)} } & \makecell{ 0.186 {\footnotesize(0.184, 0.187)} } & \makecell{ \underline{ 0.99 } {\footnotesize(0.983, 0.998)} } & 0.447 & 0.279 \\
LINEARITY & \makecell{ 0.683 {\footnotesize(0.679, 0.687)} } & \makecell{ 0.447 {\footnotesize(0.444, 0.450)} } & \makecell{ 0.267 {\footnotesize(0.263, 0.272)} } & 0.780 & 0.683 \\
VSFA & \makecell{ 0.899 {\footnotesize(0.891, 0.907)} } & \makecell{ 0.611 {\footnotesize(0.606, 0.617)} } & \makecell{ 0.659 {\footnotesize(0.650, 0.667)} } & 0.739 & 0.899 \\
PAQ-2-PIQ & \makecell{ 0.521 {\footnotesize(0.518, 0.524)} } & \makecell{ 0.341 {\footnotesize(0.338, 0.343)} } & \makecell{ 0.449 {\footnotesize(0.443, 0.454)} } & 0.675 & 0.521 \\
SPAQ & \makecell{ 0.671 {\footnotesize(0.665, 0.678)} } & \makecell{ 0.536 {\footnotesize(0.531, 0.542)} } & \makecell{ 0.493 {\footnotesize(0.488, 0.499)} } & 0.637 & 0.671 \\
TRES & \makecell{ 0.433 {\footnotesize(0.431, 0.435)} } & \makecell{ 0.305 {\footnotesize(0.304, 0.307)} } & \makecell{ 0.320 {\footnotesize(0.317, 0.323)} } & 0.627 & 0.433 \\
MANIQA & \makecell{ \textbf{ 0.104 } {\footnotesize(0.101, 0.107)} } & \makecell{ \textbf{ 0.078 } {\footnotesize(0.076, 0.08)} } & \makecell{ 0.986 {\footnotesize(0.979, 0.993)} } & \bfseries 0.207 & \underline {{0.175}} \\
\bottomrule
\end{tabular}
\caption{Metrics' robustness to tested adversarial attacks according to different stability measures. The results for abs. gain, rel. gain and R-score were averaged across different types of attacks and test datasets, so they are presented with confidence intervals. The $E_{score}$ and $W_{score}$ were calculated using the whole set of attacked results without averaging.}
\label{tab:averages_scores}
\end{table*}

\begin{figure}[tb]
  \centering
  \includegraphics[width=0.98\columnwidth]{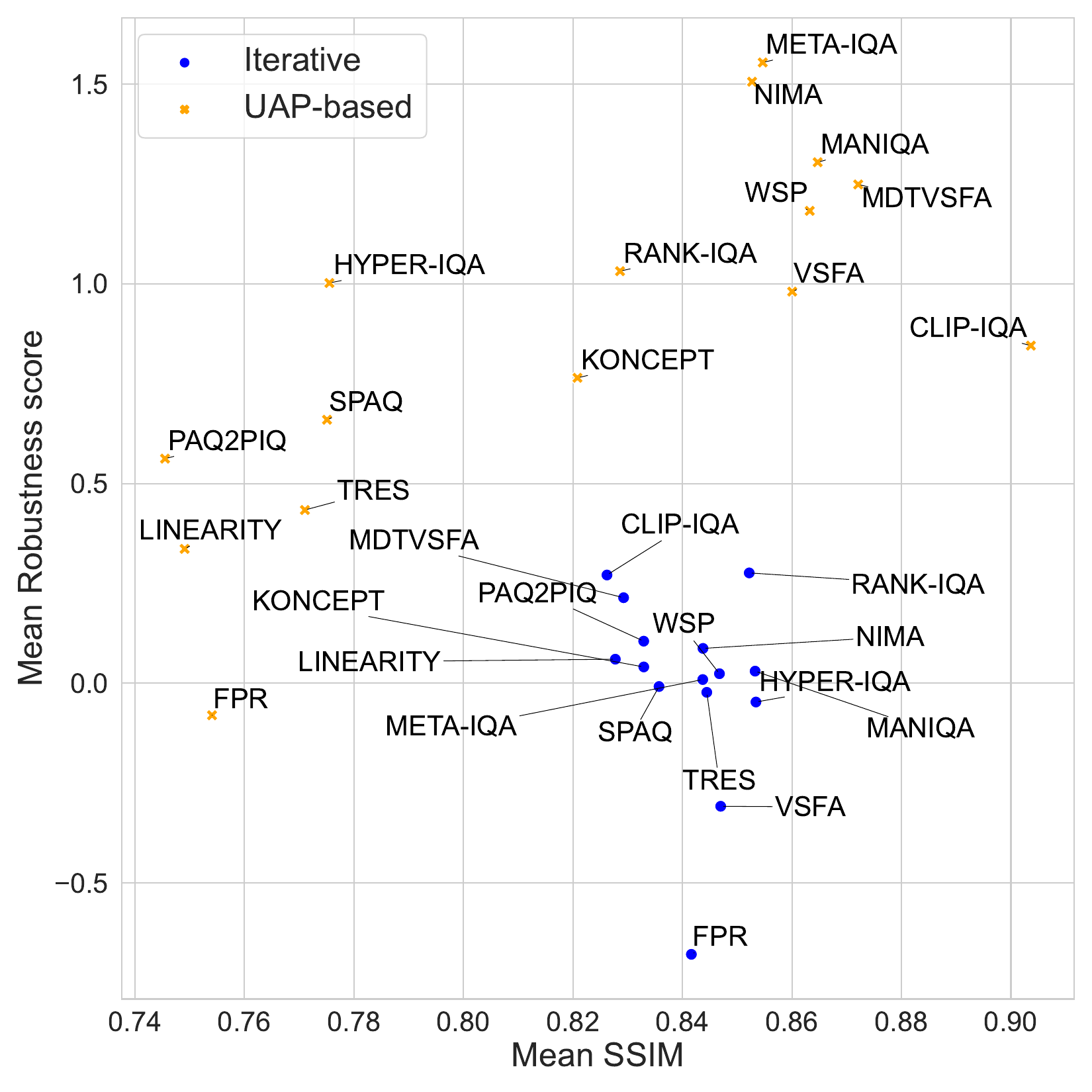}
  \caption{
  Mean robustness score of compared metrics versus SSIM averages for UAP-based and iterative attacks.
    }
  
  \label{fig:uap_iterative_aggregated_mean_rscore_ssim}
\end{figure}

\textbf{Overall metrics' robustness comparison.}
Table~\ref{tab:averages_scores} and Fig.~\ref{fig:uap_iterative_aggregated_mean_rscore_ssim} show the general results of our study. First, we see that iterative attacks are more efficient against all metrics. However, metrics' robustness is different for UAP and iterative attacks. We summarised the robustness of all attack types in the table and compared them using various measures. According to absolute and relative gain, the leaders are the same: MANIQA, NIMA and RANK-IQA, and they also perform well based on other measures. META-IQA and MDTVSFA have high robustness scores. Energy measures also show similar results. FPR is the least stable to adversarial attacks, considering all tests and measures. 

\textbf{One-sided Wilcoxon signed-rank tests.}
To study the statistical difference in the results, we conducted one-sided Wilcoxon tests on the values of absolute gains for all pairs of metrics. A table with detailed test results for different types of attacks can be found in the supplementary materials \cite{antsiferova2023comparing}. All metrics are statistically superior to the FPR metric, which means that FPR can be significantly increased under the influence of any of the considered attacks. MANIQA, on the contrary, turns out to be one of the most stable metrics for all attacks on average, but it is inferior to CLIP-IQA on FGSM-based attacks. 
Overall, the results of the Wilcoxon one-sided tests are consistent with our evaluations of the obtained results.

\textbf{Stable metrics feature analysis.}
To analyze the relationship of metrics' architectures with robustness, we summarised the main features of tested metrics in Table 1 of the supplementary materials. A common feature of robust metrics is the usage of the input image cropping or resizing. High stability to attacks was also shown by META-IQA, which does not transform input images but uses a relatively small backbone network that leverages prior knowledge of various image distortions obtained during so-called meta-learning.

\section{Conclusion}
\label{conclusion}
This paper analyzed the robustness of 15 no-reference image/video-quality metrics to different adversarial attacks. Our analysis showed that all metrics are susceptible to adversarial attacks, but some are more robust than others. MANIQA, META-IQA, NIMA, RANK-IQA and MDTVSFA showed high resistance to adversarial attacks, making their usage in practical applications safer than other metrics. We published this comparison online and are accepting new metrics submissions. This benchmark can be helpful for researchers and companies who want to make their metrics more robust to potential attacks. 

In this paper, we revealed ways of cheating on image quality measures, which can be considered to have a potential negative social impact. However, as was discussed in the Introduction, the vulnerabilities of image- and video-quality metrics are already being exploited in some real-life applications. At the same time, only a few studies have been published. We open our findings to the research community to increase the trustworthiness of image/video processing and compression benchmarks. Limitations of our study are listed in the supplementary materials \cite{antsiferova2023comparing}.

\appendix

\section{Acknowledgments}
The authors would like to thank the video group of MSU Graphics and Media Laboratory, especially Kirill Malyshev and Vyacheslav Napadovsky, for setting up the infrastructure and helping to receive computational results for this research.
The work was supported by a grant for research centers in the field of artificial intelligence, provided by the Analytical Center in accordance with the subsidy agreement (agreement identifier 000000D730321P5Q0002) and the agreement with the Ivannikov Institute for System Programming dated November 2, 2021 No. 70-2021-00142.

\bibliography{aaai24}

\ifarXiv
    \foreach \x in {1,...,\numbersupplementpages}
    {
        \includepdf[pages={\x}]{\supplementfilename}
    }
\fi

\end{document}